\documentclass[runningheads, orivec]{llncs}
\usepackage[T1]{fontenc}
\usepackage[numbers]{natbib}
\usepackage{graphicx, hyperref, amsmath, amssymb, bbm}
\usepackage{color}

\usepackage{adjustbox}
\graphicspath{{figurer/}}
\usepackage{booktabs, multirow}  
\usepackage{lmodern}
\usepackage{siunitx}
\usepackage{etoolbox}
\usepackage{enumitem}

\renewrobustcmd{\bfseries}{\fontseries{b}\selectfont}
\renewrobustcmd{\boldmath}{}

\usepackage{xcolor,colortbl}  
\definecolor{col_ind}{HTML}{E81A20}
\definecolor{col_emp}{HTML}{FF7F00}
\definecolor{col_par}{HTML}{FFDF0F}
\definecolor{col_gen}{HTML}{33A02C}
\definecolor{col_sep}{HTML}{A6CEE3}
\definecolor{col_sur}{HTML}{1F78B4}

\usepackage{accents}
\newcommand{\thickbar}{} 
\DeclareRobustCommand*\thickbar[1]{\accentset{\rule{.35em}{.65pt}}{#1}}

\newcommand{\reals}{\mathbb{R}}

\DeclareMathOperator*{\argmin}{arg\,min}

\newcommand{\E}{\mathbb{E}}

\newcommand{\given}{\,|\,}

\newcommand{\x}{\boldsymbol{x}}

\newcommand{\diff}{\mathop{}\!\mathrm{d}}

\newcommand{\s}{{\mathcal{S}}}

\newcommand{\pow}{{\mathcal{P}}}

\newcommand{\bSigma}{{\boldsymbol{\Sigma}}}

\newcommand{\gam}{\text{gam}}

\newcommand{\more}{\text{more}}

\newcommand{\Sbar}{\thickbar{\mathcal{S}}}

\begin{document}
\title{\texorpdfstring{Computing Conditional Shapley Values Using\newline Tabular Foundation Models}{Computing Conditional Shapley Values Using Tabular Foundation Models}}
%
%
\author{Lars Henry Berge Olsen\inst{1}\orcidID{0009-0006-9360-6993} \and \newline
Dennis Christensen\inst{2}\orcidID{0000-0002-7540-7695}}
\authorrunning{L. H. B. Olsen and D. Christensen}
%
\institute{Norwegian Computing Center \email{lhbolsen@nr.no} \and
Norwegian Defence Research Establishment (FFI)
\email{dennis.christensen@ffi.no}}
\maketitle              
\begin{abstract}
    Shapley values have become a cornerstone of explainable AI, but they are computationally expensive to use, especially when features are dependent. Evaluating them requires approximating a large number of conditional expectations, either via Monte Carlo integration or regression. Until recently it has not been possible to fully exploit deep learning for the regression approach, because retraining for each conditional expectation takes too long. Tabular foundation models such as TabPFN overcome this computational hurdle by leveraging in-context learning, so each conditional expectation can be approximated without any re-training. In this paper, we compute Shapley values with multiple variants of TabPFN and compare their performance with state-of-the-art methods on both simulated and real datasets. In most cases, TabPFN yields the best performance; where it does not, it is only marginally worse than the best method, at a fraction of the runtime. We discuss further improvements and how tabular foundation models can be better adapted specifically for conditional Shapley value estimation.
\end{abstract}
\section{Introduction}
In the age of big data, large machine learning (ML) models are being applied throughout modern science, with applications in image recognition \cite{krizhevsky2012imagenet}, cancer prognosis \cite{kourou2015machine} and protein structure prediction \cite{jumper2021highly}. Unlike classical statistical models like linear or logistic regression, these ML models typically contain millions (or billions) of parameters, making them virtually impossible for humans to interpret directly. This lack of transparency has led to the development of the field of explainable AI (XAI), which devises algorithms for interpreting ML model parameters and ranking features by their importance. Interest in XAI has surged following legislation by the European Union, such as the General Data Protection Regulation (GDPR) and the Artificial Intelligence Act, both of which introduce transparency and explanation requirements for automated decision‑making \cite{EU2024AI, EU2016GDPR}.

One of the most successful approaches to XAI is the use of Shapley values \cite{shapley1953value}, which originate in cooperative game theory but have been used extensively to explain ML predictions \cite{aas2021explaining, lundberg2017unified, olsen2022using, strumbelj2014explaining}. Shapley values have a sound mathematical foundation and are entirely model-agnostic, which means they can be used to explain virtually any supervised learning model. Although they are relatively straightforward to define, they are computationally challenging to evaluate, particularly when the features are dependent, as is the case for most real-world problems.

To compute the Shapley value of a feature in a predictive model, a large number of conditional expectations have to be evaluated. Over the past decade, multiple approximation schemes for these conditional expectations have been proposed, varying widely in flexibility and computation time (see \cite{olsen2024comparative} for an extensive comparative study). Broadly speaking, these can be divided into two categories based on whether they compute the expectations via (i) Monte Carlo integration or (ii) least-squares regression. Of the former category (i), nonparametric generative approaches such as conditional inference trees \cite{redelmeier2020explaining} and variational autoencoders \cite{olsen2022using} tend to outperform parametric modeling. The latter category (ii) can be split into two subcategories: in the first, each conditional expectation is estimated by its own separately trained regression model, whereas in the second, a single model estimates all conditional expectations. The comparison in \cite{olsen2024comparative} shows that separately trained models quite consistently yield the best performance. However, because each conditional expectation requires its own trained model, the set of feasible models for separate regression has so far been limited by time constraints. In particular, deep learning---the best performing surrogate approach in \cite{olsen2024comparative}--- has so far been impractical to use for separate fitting.\footnote{The preprint version of \cite{olsen2024comparative} (see \cite{olsen2023comparative}, Appendix C.3.14) does consider separately trained fully connected feedforward neural networks with 3 layers, but these do not yield competitive performance.}

The last year has seen the advent of the first successful foundation model for moderately sized tabular data, namely the tabular prior fitted network (TabPFN) and its variants \cite{hollmann2025accurate}. With a transformer-based model architecture, TabPFN is pre-trained on more than 100 million synthetic datasets and can be applied to generic regression and classification tasks without further training using in-context learning. It has been demonstrated to outperform tree-based models such as XGBoost and CatBoost on multiple standard regression benchmarks at only a fraction of the runtime. This makes TabPFN a natural and indeed promising option for computing Shapley values.

In this paper, we explore the computation of Shapley values in XAI using TabPFN. To our knowledge, prior work has only exploited TabPFN’s architecture to improve on the KernelSHAP algorithm when explaining predictions made by TabPFN itself \cite{muschalik2024shapiq, rundel2024interpretable}. In contrast, we use TabPFN for model‑agnostic explanations, allowing us to explain predictions from any predictive model rather than just TabPFN. We extend the comparative study from \cite{olsen2024comparative}, and evaluate multiple variants of TabPFN in a simulation study and on the five benchmark datasets, under both the separate and surrogate regression paradigms. Our results show that TabPFN outperforms the existing state-of-the-art baseline models in many cases, particularly when the predictive model is smooth. For non-smooth predictors, TabPFN performs only marginally worse than tree-based alternatives, but at a fraction of the computation time. In summary, this work clearly demonstrates the potential of computing conditional Shapley values using tabular foundation models. 

\section{Shapley values}\label{sec:shapley}
Shapley values \cite{shapley1953value} originate from cooperative game theory and provide a method for distributing the total value of a game fairly among its $M$ players. More specifically, let $v : \mathcal{P}(\mathcal{M}) \to \mathbb{R}$ be a \emph{contribution function}, meaning a map from the power set of $\mathcal{M} = \{1,\dots,M\}$ to the real numbers. The elements of $\mathcal{P}(\mathcal{M})$, i.e.,~the subsets of $\mathcal{M}$, are typically called \emph{coalitions}. Loosely speaking, we think of $v(\mathcal{S})$ as the payoff ascribed to the players in coalition $S$ if they decide to cooperate in the game $v$. To distribute the total payout of the game among its players, we seek to assign to each player $j\in\mathcal{M}$ a value $\phi_j(v)$, satisfying the following four axioms.
\begin{itemize}
    \item \textit{Efficiency:} $\sum_{j=1}^M\phi_j(v) = v(\mathcal{M}) - v(\emptyset)$, where $\emptyset$ is the empty set.

    \item \textit{Symmetry:} For all $j, k \in \mathcal{M}$, if $v(\mathcal{S}\cup\{j\}) = v(\mathcal{S}\cup\{k\})$ for all $\mathcal{S}\in\mathcal{P}(\mathcal{M})$, then $\phi_j(v) = \phi_k(v)$.

    \item \textit{Dummy:} For all $j\in\mathcal{M}$, if $v(\mathcal{S}\cup\{j\}) = v(\mathcal{S})$ for all $\mathcal{S}\in\mathcal{P}(\mathcal{M})$, then $\phi_j(v) = 0$.

    \item \textit{Linearity:} If $\{v_1, \dots, v_n\}$ are $n$ games on the same set $\mathcal{M}$, $c_1, \dots, c_n \in \mathbb{R}$ are constants, and $v(\mathcal{S}) = \sum_{i=1}^n c_i v_i(\mathcal{S})$, then for all $j\in\mathcal{M}$, $\phi_j(v) = \sum_{i=1}^n c_i \phi_j(v_i)$.
\end{itemize}
Shapley~\cite{shapley1953value} proved that there exists a unique solution satisfying all four axioms, given in closed form by
\begin{align}\label{eq:shapley}
    \phi_j(v)
    = \!\!\sum_{\mathcal{S} \subseteq \mathcal{M}\setminus\{j\}}
    \frac{|\mathcal{S}|!(M-|\mathcal{S}|-1)!}{M!}
    \big( v(\mathcal{S} \cup \{j\}) - v(\mathcal{S}) \big),
\end{align}
where $|\mathcal{S}|$ denotes the cardinality of the coalition $\mathcal{S}$. Due to uniqueness, we refer to $\{\phi_1(v), \dots, \phi_M(v)\}$ as \emph{the} Shapley values of the game $v$. We will also write $\phi_j$ for $\phi_j(v)$ when the game $v$ is implied from context.

In supervised learning, Shapley values are used as a model-agnostic explanation tool \cite{lundberg2017unified, strumbelj2014explaining}. Here, the predictive model $f(\boldsymbol{x})$ plays the role of the game, and the $M$ features play the role of the players. For a particular instance $\boldsymbol{x}^*$, the Shapley values $\{\phi_1^*, \dots, \phi_M^*\}$ satisfy $f(\boldsymbol{x}^*) = \phi_0 + \sum_{j=1}^M \phi_j^*$, where $\phi_0 = \mathbb{E}[f(\boldsymbol{x})]$. Thus, the Shapley values $\phi_j^*$ describe the contribution of each feature $j$ to the deviation of $f(\boldsymbol{x}^*)$ from the global average prediction $\phi_0$.

Adapting Shapley values to model explanations requires the specification of a suitable contribution function $v$, where $v(\mathcal{S})$ should specify the value of the model when only the features from coalition $\mathcal{S}$ is known. Here, we follow the convention of Lundberg and Lee~\cite{lundberg2017unified}, using the conditional expectation
\begin{multline}
\label{eq:contrib}
    v(\mathcal{S}) 
    =
    \E\left[ f(\boldsymbol{x})\given\boldsymbol{x}_{\mathcal{S}} = \boldsymbol{x}_{\mathcal{S}}^* \right] 
    =
    \E\left[ f(\boldsymbol{x}_{\mathcal{S}}, \boldsymbol{x}_{\Sbar})\given\boldsymbol{x}_{\mathcal{S}} = \boldsymbol{x}_{\mathcal{S}}^* \right] \\ =
    \int f(\boldsymbol{x}_{\mathcal{S}}, \boldsymbol{x}_{\Sbar}^*) p(\boldsymbol{x}_{\thickbar{\mathcal{S}}}\given\boldsymbol{x}_{\mathcal{S}} = \boldsymbol{x}_{\mathcal{S}}^*) \diff \boldsymbol{x}_{\protect\thickbar{\mathcal{S}}},
\end{multline}
where $\boldsymbol{x}_{\mathcal{S}} = \{x_j : j\in\mathcal{S}\}$ and $\thickbar{\mathcal{S}} = \mathcal{M}\setminus\mathcal{S} = \{j\in\mathcal{M} : j \notin \mathcal{S}\}$. In other words, $v(\mathcal{S})$ is the expected prediction given that the features in $\mathcal{S}$ are fixed to their observed values $\boldsymbol{x}_{\mathcal{S}}^*$.
This formulation is referred to as \textit{conditional Shapley values} and has become standard in Shapley value-based explainability. The alternative, marginal Shapley values \cite{aas2021explaining, covert2021explaining}, will not be considered here.

Note that~\eqref{eq:shapley} averages the marginal contribution of player $j$ over all possible coalitions, but its exact computation requires $2^M$ evaluations and therefore grows exponentially in the number of players. For high-dimensional problems where this is not feasible, there exist Monte Carlo-based estimators which bypass this computational hurdle \cite{aas2021explaining, covert2021improving, lundberg2017unified, olsen2025improving}. However, since this paper concerns ways of computing $v(\mathcal{S})$ and we want to avoid the additional variance added by these Monte Carlo methods, we restrict ourselves to lower-dimensional datasets for which exact computation of the $2^M$ coalitions is possible in reasonable time.

Because the conditional distribution $p(\boldsymbol{x}_{\Sbar}\given\boldsymbol{x}_{\mathcal{S}}=\boldsymbol{x}_{\mathcal{S}}^*)$ is typically unknown and must be approximated, computing Shapley values relies on estimating \eqref{eq:contrib} across the coalitions $\mathcal{S}$. To this end, two broad paradigms have emerged: (i) Monte Carlo estimation and (ii) regression-based estimation. Both are widely used in modern XAI and come with trade-offs between accuracy and computational cost. We provide a brief overview of the two approaches here and refer to \cite{olsen2024comparative} for a more comprehensive account.

\subsection{Monte Carlo estimation}
The Monte Carlo paradigm approximates the conditional expectation directly by drawing samples from the conditional distribution:
\begin{align}
    v(\mathcal{S}) = \E\left[f(\boldsymbol{x}_{\mathcal{S}}, \boldsymbol{x}_{\thickbar{\mathcal{S}}})\given\boldsymbol{x}_{\mathcal{S}} = \boldsymbol{x}_{\mathcal{S}}^* \right] 
    \approx
    \frac{1}{K} \sum_{k=1}^K f(\boldsymbol{x}_{\mathcal{S}}^*, \boldsymbol{x}_{\thickbar{\mathcal{S}}}^{[k]}),
\end{align}
where $\boldsymbol{x}_{\thickbar{\mathcal{S}}}^{[k]} \sim p(\boldsymbol{x}_{\thickbar{\mathcal{S}}}\given\boldsymbol{x}_{\mathcal{S}} = \boldsymbol{x}_{\mathcal{S}}^*)$ independently for $k=1, \dots, K$. This approach is conceptually straightforward and imposes no constraints on the form of the predictive model $f$. Obtaining realistic conditional samples is often the main bottleneck. In~\cite{olsen2024comparative}, the Monte Carlo-based estimators are grouped into four method classes based on how they estimate the conditional distribution $p(\boldsymbol{x}_{\thickbar{\mathcal{S}}}\given \boldsymbol{x}_{\mathcal{S}}=\boldsymbol{x}^*_{\mathcal{S}})$:
\begin{enumerate}[label=(\alph*)]
    \item The \textit{independence method} assumes the features to be mutually independent, so the conditional distribution reduces to the marginal $p(\boldsymbol{x}_{\thickbar{\mathcal{S}}})$. This makes sampling simple but may break important dependencies in the data. 
    \item The \textit{empirical method} draws samples from training observations that resemble $\boldsymbol{x}^*_{\mathcal{S}}$ in the conditioning features, preserving some of the dependence structure without having to model the full joint distribution. 
    \item \textit{Parametric methods} assume a specific parametric form for the joint distribution allowing closed-form conditional sampling, such as multivariate Gaussian or copula. Their accuracy depends on how well the parametric model fits the true data distribution. 
    \item \textit{Generative methods} also learn the conditional distribution but without relying on parametric assumptions.  Flexible ML models such as conditional inference trees or variational autoencoders are used to approximate the conditional density and generate samples.
\end{enumerate}

\subsection{Regression-Based Estimation}
Rather than estimating the conditional distribution $p(\boldsymbol{x}_{\thickbar{\mathcal{S}}}\given\boldsymbol{x}_{\mathcal{S}} = \boldsymbol{x}_{\mathcal{S}}^*)$ directly, regression-based estimation uses that conditional expectation in \eqref{eq:contrib} minimizes the mean squared error
\begin{align}\label{eq:regression}
    v(\mathcal{S}) = \E\left[f(\boldsymbol{x}_{\mathcal{S}}, \boldsymbol{x}_{\thickbar{\mathcal{S}}})\given\boldsymbol{x}_{\mathcal{S}} = \boldsymbol{x}_{\mathcal{S}}^* \right] = \argmin_{c\in\reals}
    \mathbb{E}\!\left[
        (f(\boldsymbol{x}) - c)^2
        \given 
        \boldsymbol{x}_{\mathcal{S}} = \boldsymbol{x}_{\mathcal{S}}^*
    \right].
\end{align}
Thus, any regression model trained to predict $f(\boldsymbol{x})$ from the variables in $\mathcal{S}$ under squared loss provides an estimator of $v(\mathcal{S})$.

Regression-based estimation ban be subdivided further into two approaches. The first is is to fit \textit{separate regression} models $g_{\mathcal{S}}(\boldsymbol{x}_{\mathcal{S}})$ for each coalition $\mathcal{S}$. This offers considerable flexibility, since each model can be tailored to the specific conditioning set, but it becomes computationally expensive as the number of features grows and the number of coalitions increases.  An alternative is \textit{surrogate regression},\footnote{Although the term \emph{amortized regression} is possibly more accurate, we use the same terminology as in \cite{olsen2024comparative} for consistency.} where instead a single model $g(\tilde{\boldsymbol{x}})$ is trained. This model takes in a fixed-length augmented input $\tilde{\boldsymbol{x}}$ which, for a given coalition $\mathcal{S}$, only contains the observed feature values $\boldsymbol{x}_{\mathcal{S}}$ (the unknown features $\boldsymbol{x}_{\thickbar{\mathcal{S}}}$ are zeroed), along with binary indicators specifying the coalition $\mathcal{S}$. For example, if $M=3$, then any observation $\boldsymbol{x}$ is replaced by the six observations
\begin{align}\label{eq:mask}
    \tilde{\boldsymbol{x}} = \begin{bmatrix} \tilde{x}_1 \\ \tilde{x}_2 \\ \tilde{x}_3 \\ \tilde{x}_{1,2} \\ \tilde{x}_{1,3} \\ \tilde{x}_{2,3}\end{bmatrix} = 
    \begin{bmatrix}
    x_{1} & 0 & 0 & 0 & 1 & 1 \\
    0 & x_{2} & 0 & 1 & 0 & 1 \\
    0 & 0 & x_{3} & 1 & 1 & 0 \\
    x_{1} & x_{2} & 0 & 0 & 0 & 1 \\
    x_{1} & 0 & x_{3} & 0 & 1 & 0 \\
    0 & x_{2} & x_{3} & 1 & 0 & 0 \\
    \end{bmatrix},
\end{align}
and the corresponding target $f(\boldsymbol{x})$ is duplicated six times. The surrogate model thus learns to approximate the contribution functions for all subsets simultaneously, substantially improving scalability while still providing accurate estimates of the conditional expectations. However, the size of the augmented dataset grows exponentially with $M$, making it intractable for higher-dimensional problems. A possible solution to this is to only sample a large subset of rows from the augmented dataset.

\section{Computing Shapley values with TabPFN}
Introduced by Hollmann et al.~\cite{hollmann2025accurate}, TabPFN is a tabular foundation model that achieves state-of-the-art performance on a wide range of regression and classification tasks for small- to medium-sized datasets. Traditionally, supervised models are first trained on a dataset $(\boldsymbol{X}_{\text{train}}, y_{\text{train}}) = \{(\boldsymbol{x}_{\text{train}}^{[i]}, y_{\text{train}}^{[i]})\}_{i=1}^{N_{\text{train}}}$ and then used to make predictions for new observations $\boldsymbol{X}_{\text{test}} = \{(\boldsymbol{x}_{\text{test}}^{[i]})\}_{i=1}^{N_\text{test}}$. TabPFN, on the other hand, takes the tuple $(\boldsymbol{X}_{\text{train}}, y_{\text{train}}, \boldsymbol{X}_{\text{test}})$ as input and makes predictions for $\boldsymbol{X}_{\text{test}}$ in a single forward pass. The model was pre-trained on more than 100 million synthetic datasets sampled from a prior distribution explicitly designed to generate diverse tabular problems. When making predictions for $\boldsymbol{X}_{\text{test}}$, the model leverages in-context-learning \cite{brown2020language}, using the training data $(\boldsymbol{X}_{\text{train}}, y_{\text{train}})$ as context. Hence, TabPFN can be used repeatedly without any retraining or hyperparameter tuning. Given its superb performance, this makes it a highly attractive option for approximating conditional Shapley values, both in the separate and surrogate regression schemes.

In our comparative study, we compare both TabPFN versions 2 and 2.5 (release v6.3.0). Version 2 is publicly available and was pre‑trained solely on synthetic datasets. Version 2.5 is available for non‑commercial use and comes in multiple variants, two of which are considered here: one pre-trained solely on synthetic data (the default choice) and another additionally trained on real data. We refer to these as v2.5-D (default) and v2.5-R (real), respectively. It is possible that some of the datasets we consider in this work were included in the pre-training of TabPFN~v2.5-R, but even if that is the case, the model was trained to accurately predict the responses, not to solve the regression problem~\eqref{eq:regression}. Therefore, we consider the effects of any data leakage to be minor. By default, TabPFN produces predictions by averaging an ensemble obtained from multiple forward passes with slightly altered inputs (see \cite{hollmann2025accurate} for details). We also compare using a single estimator (no ensembling) with the default ensemble of eight estimators.

In the separate regression approach, we apply TabPFN independently to each coalition $\mathcal{S}$. That is, we only use the features in $\mathcal{S}$ in the in-context learning phase. The surrogate regression approach is further subdivided into two categories, which we refer to as the direct and augmented surrogate approaches, respectively. In the direct approach, the full dataset is used as context, and TabPFN makes predictions for a coalition $\mathcal{S}$ simply by treating all features in $\thickbar{\mathcal{S}}$ as missing using TabPFN's built-in handling of missing values. In the augmented approach, we use the full dataset as context after modifying it so that each coalition produces a distinct block of missing features. For each coalition $\mathcal{S}$ we sample $L$ training instances from the original dataset, and mask all features in $\thickbar{\mathcal{S}}$ as missing by setting them to \texttt{NaN}, which TabPFN handles internally. This corresponds to~\eqref{eq:mask}, but with \texttt{NaN} instead of zeros in the first three columns. We choose $L$ such that $(2^M - 2)L = 5 \times 10^4$, corresponding to the upper limit for TabPFN~v2.5's training without disabling the built-in restrictions. We subtract 2 from $2^M$ because $v(\emptyset)$ and $v(\mathcal{M})$ are trivial to evaluate. Note that increasing the size of the data in this way slows down the predictions, since the context is now much larger than the original dataset.

Finally, because the augmented surrogate approach yields relatively few examples for each coalition, we also consider applying one TabPFN model per coalition size rather than per coalition. We refer to this as the surrogate augmented coalition approach. In this case the required amount of augmented data satisfies $\binom{M}{|\mathcal{S}|}L = 5 \times 10^4$, allowing substantially more context learning examples for each model while remaining within the training limits of TabPFN~v2.5. 

\section{Simulation Study}
\label{sec:simulation}
We first compare the approaches in a simulation study where the data are generated from a known multivariate Gaussian distribution. This ensures that the conditional distributions $p_{\text{true}}(\boldsymbol{x}_{\thickbar{\mathcal{S}}}\given \boldsymbol{x}_{\mathcal{S}})$ are analytically tractable, and so the true Shapley values ${\phi}_{\text{true}}$ can be computed with arbitrary precision. To generate the response variable, we employ the most challenging experiment from \cite[Section~4]{olsen2024comparative}, the \texttt{gam\_more\_interactions} procedure. This combines nonlinear additive effects with pairwise nonlinear interaction terms, providing a demanding and representative test case. Following \cite{olsen2024comparative}, we generate $M=8$ features by sampling
\begin{align*}
    \x^{[i]} \sim \mathcal{N}_8(\mathbf{0}, \bSigma), 
    \qquad 
    \bSigma_{ij} = \rho^{|i-j|}, \qquad 
    \rho \in \{0, 0.3, 0.5, 0.9\},
\end{align*}
and use $N_{\text{train}} = 1000$ observations for model training and $N_{\text{test}} = 250$ for explanation. The function \texttt{gam\_more\_interactions} is defined as
\begin{align*}
    f_{\gam\_\more}(\x) 
    = \beta_0 + \sum_{j=1}^{M}\beta_j \cos(x_j)
    + \gamma_1 g(x_1, x_2) + \gamma_2 g(x_3, x_4),
\end{align*}
where $g(x_j, x_k) = x_j x_k + x_j x_k^2 + x_k x_j^2$, and we use the same coefficient vectors 
$\boldsymbol{\beta} = \{1.0, 0.2, -0.8, 1.0, 0.5, -0.8, 0.6, -0.7, -0.6\}$ and 
$\boldsymbol{\gamma} = \{0.8, -1.0, -2.0, 1.5\}$ as in \cite{olsen2024comparative}. The observed response is given by $y^{[i]} = f_{\gam\_\more}(\x^{[i]}) + \varepsilon^{[i]}$ with $\varepsilon^{[i]} \sim \mathcal{N}(0,1)$, independently. As predictive model $f$, we use a generalized additive model fitted with splines for the nonlinear main effects and tensor-product smooths for the interaction terms. To assess the accuracy of the estimation methods, we compute the mean absolute errors ($\operatorname{MAE}$)
\begin{align}\label{eq:MAE}
    \operatorname{MAE}^{[i]}
    =
    \frac{1}{M} \sum_{j=1}^M
    \left| \phi_{\text{true}, j}(\x^{[i]}) - \widehat{\phi}_{j}(\x^{[i]}) \right|
\end{align}
between the estimated Shapley values $\widehat{\phi}$ and $\phi_{\text{true}}$ for each $\boldsymbol{x}^{[i]}$ in the test set, a standard metric in the literature \cite{aas2021explaining, olsen2024comparative, olsen2025improving, redelmeier2020explaining}.

\begin{figure}
    \centering
    \includegraphics[width=1\linewidth]{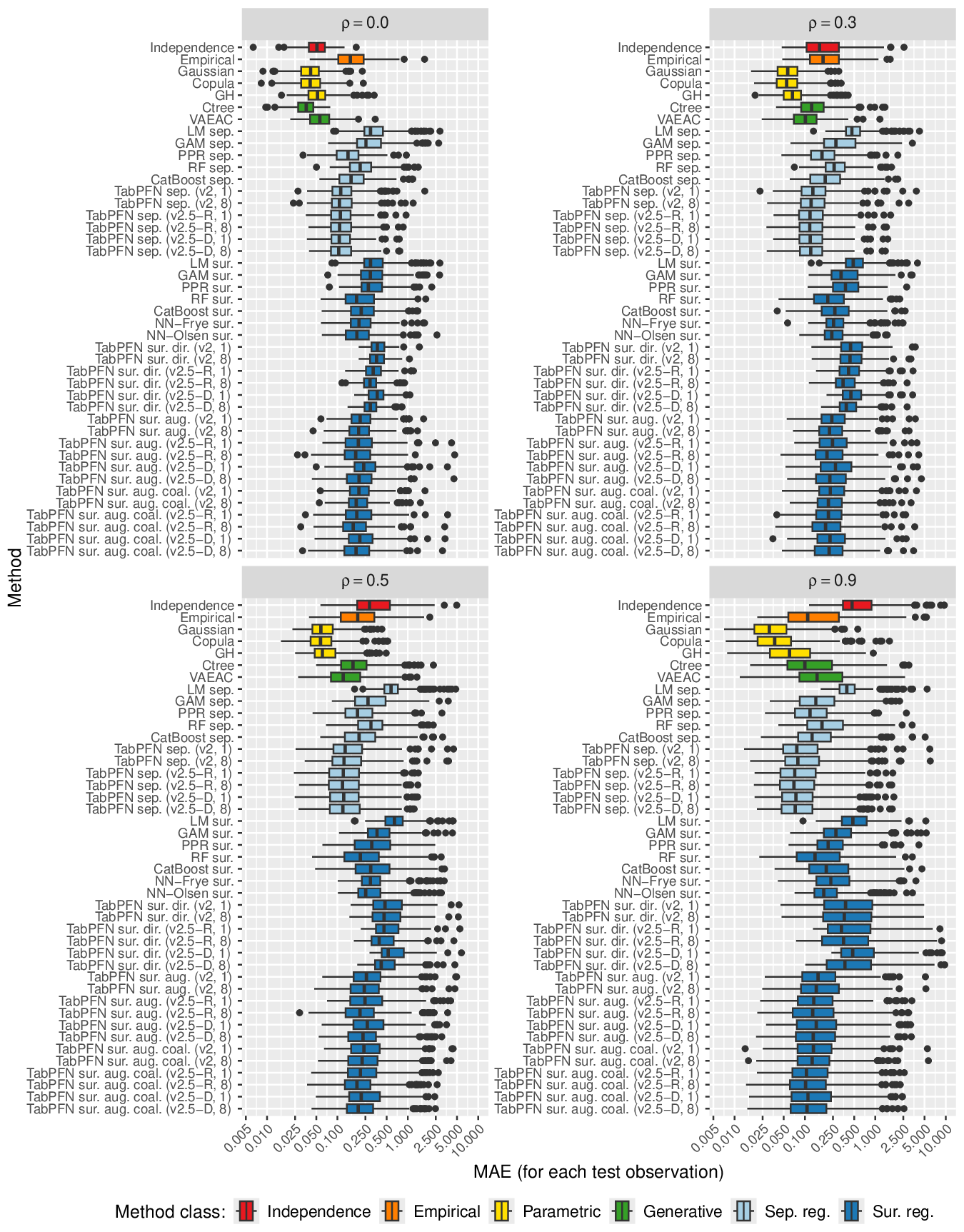}
    \label{fig:simulation_study}
    \caption{{\small 
    Results of the \texttt{gam\textunderscore more\textunderscore interactions} experiment: Boxplots of the mean absolute error between the true and estimated Shapley values for the test observations using different methods and for different dependence levels $\rho$. Abbreviations: GH = generalized hyperbolic, Ctree = conditional inference tree, VAEAC = variational autoencoders with arbitrary conditioning, LM = linear regression model, GAM = generalized additive model, PPR = projection pursuit regression, RF = random forest, NN = neural network. For a complete description of all models and appropriate references, see~\cite{olsen2024comparative}.
    }}
\end{figure}
In Fig.~\ref{fig:simulation_study}, we show the MAE for each test observation. The separate regression methods improve as $\rho$ increases, although the parametric methods remain superior. However, they benefit from an inherent advantage: they assume Gaussian or near‑Gaussian data, which matches the true data-generating process. We also repeated the experiment replacing the Gaussian with the Burr distribution, in which case the performance of the parametric models deteriorated notably.

After the separate regression class, the generative methods form the next-best class for $\rho \in \{0, 0.3\}$, but for $\rho = 0.3$ both the separate and surrogate classes with TabPFN are close, and they outperform the generative class for $\rho \in \{0.5, 0.9\}$. The simpler separate regression approaches fall behind TabPFN because they lack the complexity needed to capture the nonlinear interaction terms. In fact, TabPFN is the best-performing method within both the separate and surrogate regression classes.

Within the separate regression class, there are no substantial performance differences between TabPFN v2, v2.5‑D, and v2.5‑R, although v2 displays a few more pronounced outliers for higher values of $\rho$. Using one estimator instead of eight offers no meaningful loss in performance, while reducing the prediction time by a factor of eight, from 134 to 18 seconds on an Nvidia RTX 4000 GPU. The total runtime, which includes the in-context learning phase, saw a reduction from 210 to 64 seconds.

In the surrogate regression setting, the direct approach performs poorly across all versions of TabPFN. We believe that this is because the type of missing data encountered in this context, with entire columns of missing values, were not well-represented in the synthetic datasets on which TabPFN was pre-trained. In contrast, the augmented approach, which introduces missingness in the context learning phase, perform substantially better, with the variant trained on a single coalition size yielding the strongest results. This is intuitive, as this model encounters more examples for each specific missingness pattern. This makes it the best surrogate method across all settings. As before, we observe no substantial differences between the different versions of TabPFN, and little benefit from using eight estimators instead of one.

In addition to replacing Gaussian with Burr distributed data, we also explored additional experimental settings by varying the training sample size (using $N_{\text{train}} = 100$ and $N_{\text{train}} = 5000$), and replacing the predictive model $f$ with a random forest and a projection pursuit regression model. We do not include the corresponding figures here, as the overall conclusions remain similar. The most notable difference, however, is that TabPFN performs worse for the random forest, the only non-smooth predictive function considered. This is in line with the observation made by Olsen et al.~\cite{olsen2024comparative}, namely that for separate or surrogate regression, it is best to use a model which is similar to the predictive one.

\section{Real data results}
\begin{table}[t]
\caption{Summary of the real-world datasets used in our experiments. The table reports the task type, feature type, number of features $M$, number of coalitions $2^M$, average absolute correlation among continuous features (*non-numerical features are omitted), number of training and test observations and the predictive model $f$ to explain (PPR = Projection Pursuit Regression, PLS = Partial Least Squares, and RF = Random Forest).}\label{tab:real_world_data}
    \centering

    \begin{adjustbox}{width=1.0\textwidth}
    \begin{tabular}{lllcccccl}
        \toprule
         Dataset & Task Type & Features & $M$ & $2^M$ & Avg. $\lvert$Corr.$\rvert$ & $N_\text{train}$ & $N_\text{test}$ & $f$ \\
         \midrule
         \rowcolor{gray!10}
         $\texttt{Abalone}_\texttt{cont}$ & Prediction     & Continuous & \phantom{1}7 & \phantom{12}128 & $0.89$\phantom{*} & \phantom{3}3133 &            1044 & PPR \\
         $\texttt{Abalone}_\texttt{all}$  & Prediction     & Mixed      & \phantom{1}8 & \phantom{12}256 & $0.89^*$          & \phantom{3}3133 &            1044 & PPR \\
         \rowcolor{gray!10}
         \texttt{Diabetes}                     & Prediction     & Continuous &           10 & \phantom{1}1024 & $0.35$\phantom{*} & \phantom{32}332 & \phantom{1} 110 & PLS \\
         \texttt{Wine}                   & Prediction     & Continuous &           11 & \phantom{1}2048 & $0.20$\phantom{*} & \phantom{3}1349 & \phantom{1} 250 & RF \\
         \rowcolor{gray!10}
         \texttt{Adult}                 & Classification & Mixed      &           14 &           16384 & $0.06^*$          &            30000 & \phantom{1} 162 & CatBoost \\
        \specialrule{.8pt}{0pt}{6pt}
    \end{tabular}
    \end{adjustbox}
\end{table}

In this section, we compare the performance of TabPFN with other methods for computing Shapley values for five real datasets, summarized in Table~\ref{tab:real_world_data}. To ensure a fair comparison, we use the exact same \href{https://archive.ics.uci.edu}{UCI Machine Learning Repository} datasets, train/test splits and predictive models as Olsen et al.~\cite{olsen2024comparative}. This allows us to isolate the effect of our proposed TabPFN methodology without introducing variability from data or model differences. However, whereas \cite{olsen2024comparative} report CPU time, which closely approximates elapsed time when executed on a single CPU core, we run TabPFN on an Nvidia RTX 4000 GPU and therefore report elapsed time. While GPU and CPU timings are not directly comparable, using elapsed time provides a fair and interpretable measure of runtime for our experiments.

For real-world datasets, the true Shapley values are unknown, making the MAE measure~\eqref{eq:MAE} inapplicable. Instead, we use the $\operatorname{MSE}_{v}$ measure proposed by Frye et al.~\cite{frye2021shapley}, which evaluates the squared difference between the model prediction $f(\boldsymbol{x})$ and the estimated coalition value $\hat{v}(\mathcal{S}, \boldsymbol{x})$, averaged over all coalitions and test observations. That is,
\begin{align*}
    \label{eq:MSE_v}
    \operatorname{MSE}_{v} =
     \frac{1}{|\pow^*(\mathcal{M})|}\sum_{\s \in \pow^*(\mathcal{M})} \frac{1}{N_\text{test}} \sum_{i=1}^{N_\text{test}} \left( f(\boldsymbol{x}^{[i]}) - \hat{v}(\s, \boldsymbol{x}^{[i]})\right)^2\!,
\end{align*}
where $\pow^*(\mathcal{M}) = \pow(\mathcal{M})\setminus\{\emptyset, \mathcal{M}\}$. This does not require the true contribution function, and has been shown to correlate strongly with $\operatorname{MAE}$ for simulated data \cite{olsen2024comparative}.

Due to time constraints, for the \texttt{Adult} dataset we evaluated only TabPFN v2 and v2.5-D with a single estimator in the surrogate augmented and surrogate augmented coalition settings. Because these configurations did not provide competitive performance relative to separate regression, we did not pursue further evaluations of the surrogate approach here.

\begin{table}[t]
\caption{The $\operatorname{MSE}_v$ scores and times for the methods applied to the real-world datasets. See Fig.~\ref{fig:simulation_study} for abbreviations. The format of the times is days:hours:minutes:seconds, where we omit the larger units of time if they are zero, and the colors indicate the different method classes.}\label{tab:real_world_data_results}
\resizebox{\textwidth}{!}{
\begin{tabular}{lrrcrrcrrcrrcrr}
\toprule
& \multicolumn{2}{c}{\texttt{Abalone}$_\texttt{cont}$ ($M$=7)} && \multicolumn{2}{c}{\texttt{Abalone}$_\texttt{all}$ ($M$=8)} && \multicolumn{2}{c}{\texttt{Diabetes} ($M$=10)} && \multicolumn{2}{c}{\texttt{Wine} ($M$=11)} && \multicolumn{2}{c}{\texttt{Adult} ($M$=14)}  \\
\cmidrule{2-3} \cmidrule{5-6} \cmidrule{8-9} \cmidrule{11-12} \cmidrule{14-15}
\rowcolor{white} Method & $\operatorname{MSE}_{v}$ & Time & & $\operatorname{MSE}_{v}$ & Time & & $\operatorname{MSE}_{v}$ & Time & & $\operatorname{MSE}_{v}$ & Time & & $\operatorname{MSE}_{v}$ & Time \\
\specialrule{.4pt}{2pt}{0pt}
\rowcolor{col_ind!10} Independence              & 8.679 &       1:24.2 && 9.144 &       3:51.5 && 0.196 &      38.4 && 0.145 &    4:00:53.5 && 0.041 &     1:10:36.3 \\
\rowcolor{col_emp!10} Empirical                 & 1.540 &       3:43.2 &&   --- &          --- && 0.143 &      15.1 && 0.088 &    2:33:18.5 &&   --- &           --- \\
\rowcolor{col_par!10} Gaussian                  & 1.349 &       3:44.0 &&   --- &          --- && 0.127 &    2:35.1 && 0.118 &    4:08:28.6 &&   --- &           --- \\
\rowcolor{col_par!10} Copula                    & 1.223 &      15:05.5 &&   --- &          --- && 0.127 &   10:54.5 && 0.107 &    4:53:36.6 &&   --- &           --- \\
\rowcolor{col_par!10} GH                        & 1.292 &       8:39.7 &&   --- &          --- && 0.133 &    7:31.3 && 0.109 &    4:23:39.9 &&   --- &           --- \\
\rowcolor{col_par!10} Burr                      & 5.640 &       5:22.3 &&   --- &          --- &&   --- &     ---   && 0.202 &    3:52:32.7 &&   --- &           --- \\
\rowcolor{col_gen!10} Ctree                     & 1.393 &       7:40.8 && 1.424 &      19:14.1 && 0.158 &    6:46.9 && 0.102 &    1:02:41.5 &&   --- &           --- \\
\rowcolor{col_gen!10} VAEAC                     & 1.182 &    2:34:03.6 && 1.180 &   11:48:22.1 && 0.128 &   21:57.8 && 0.093 &    6:09:37.9 && 0.027 &  5:12:31:03.7 \\
\rowcolor{col_sep!10} LM sep.                   & 1.684 &          0.3 && 1.581 &          0.9 && 0.126 &       1.9 && 0.146 &          4.6 && 0.043 &     9:00:13.5 \\
\rowcolor{col_sep!10} GAM sep.                  & 1.298 &         33.7 && 1.299 &       1:09.7 && 0.126 &    1:03.6 && 0.124 &       3:24.6 && 0.033 &     2:38:52.6 \\
\rowcolor{col_sep!10} PPR sep.                  & 1.169 &       2:15.1 && 1.185 &       3:34.6 && 0.126 &    5:22.4 && 0.129 &      25:19.4 && 0.032 & 14:12:19:47.6 \\
\rowcolor{col_sep!10} RF sep.                   & 1.239 &    1:09:15.8 && 1.259 &    2:31:09.7 && 0.143 & 1:00:23.6 &&\textbf{0.071}&    9:21:12.9 && 0.027 & 98:13:33:27.4 \\
\rowcolor{col_sep!10} CatBoost sep.             & 1.190 &       6:17.1 && 1.213 &      18:25.0 && 0.135 &   18:40.6 && 0.082 &    1:41:32.7 && \textbf{0.026}& 35:09:59:59.1 \\


\rowcolor{col_sep!10} TabPFN sep.\@ (v2, 1)  & 1.137 &             33.1 &&          1.148 &          1:10.9 &&       0.126 &       4:03.8 &&   0.145 &   8:53.2 &&    0.030 &  11:15:38.0 \\
\rowcolor{col_sep!10} TabPFN sep.\@ (v2, 8)  & \textbf{1.135} &           1:54.4 &&          1.145 &          4:21.4 &&       0.125 &      12:27.4 &&   0.144 &  27:46.5 &&      0.029 &  2:23:30:09.8  \\


\rowcolor{col_sep!10} TabPFN sep.\@ (v2.5-D, 1)  & 1.138 &             32.3 &&          1.146 &          1:09.5 &&       0.126 &       4:22.6 &&   0.142 &   8:26.2 &&    0.029 &    10:44:57.6  \\
\rowcolor{col_sep!10} TabPFN sep.\@ (v2.5-D, 8)  & 1.137 &           1:52.6 &&          1.143 &          3:56.5 &&       0.125 &      13:54.0 &&   0.145 &  28:34.9 &&    0.029 &  2:21:02:51.5  \\

\rowcolor{col_sep!10} TabPFN sep.\@ (v2.5-R, 1)  & 1.138 &             34.6 &&          1.144 &          1:08.3 &&       0.126 &       4:16.3 &&   0.130 &   8:28.7 &&      0.029 &       10:48:46.0  \\
\rowcolor{col_sep!10} TabPFN sep.\@ (v2.5-R, 8)  & 1.136 &           1:51.8 &&          \textbf{1.142} &       3:55.8 &&       \textbf{0.125} &      13:41.3 &&   0.129 &  28:43.8 &&      0.029 &  2:23:03:59.4  \\

\rowcolor{col_sur!10} LM sur.                   & 2.912 &          3.7 && 2.770 &          7.2 && 0.165 &       4.4 && 0.162 &         25.9 &&   --- &           --- \\
\rowcolor{col_sur!10} GAM sur.                  & 2.611 &         41.1 && 2.557 &       1:24.6 && 0.168 &      21.3 && 0.145 &       4:01.4 &&   --- &           --- \\
\rowcolor{col_sur!10} PPR sur.                  & 1.548 &      14:57.5 && 1.538 &      55:31.6 && 0.136 &    4:13.0 && 0.149 &    1:20:15.5 &&   --- &           --- \\
\rowcolor{col_sur!10} RF sur.                   & 1.281 &    1:14:30.8 && 1.311 &    3:46:34.1 && 0.143 & 1:33:51.3 && 0.085 & 1:15:42:46.3 &&   --- &           --- \\
\rowcolor{col_sur!10} CatBoost sur.             & 1.298 &       9:10.8 && 1.348 &      29:28.6 && 0.140 &      53.4 && 0.108 &      29:06.4 &&   --- &           --- \\
\rowcolor{col_sur!10} NN-Frye sur.              & 1.244 & 3:01:48:38.7 && 1.320 & 3:16:01:07.5 && 0.154 & 3:11:39.4 && 0.170 & 1:10:53:27.4 && 0.085 &  3:16:13:50.3 \\
\rowcolor{col_sur!10} NN-Olsen sur.             & 1.169 & 3:22:23:46.0 && 1.192 & 2:01:07:30.3 && 0.135 & 1:28:54.8 && 0.130 &   23:56:47.0 && 0.045 &  3:11:33:57.9 \\

\rowcolor{col_sur!10} TabPFN sur.\@ dir.\@ (v2, 1)     & 8.878 &   16.3 && 6.741 & 1:17.3 && 0.283 & 1:01.8 && 0.237 &  2:07.7 && 0.056 & 1:13:56:40.0 \\ 
\rowcolor{col_sur!10} TabPFN sur.\@ dir.\@ (v2, 8)     & 9.000 & 1:30.6 && 6.960 & 7:53.3 && 0.285 & 7:28.1 && 0.232 & 14:37.5 && 0.052 & 7:20:31:19.2 \\ 
\rowcolor{col_sur!10} TabPFN sur.\@ dir.\@ (v2.5-D, 1) & 9.381 &   14.6 && 7.127 &   35.0 && 0.329 & 1:13.7 && 0.226 &  2:32.7 && 0.102 & 1:11:28:46.6 \\ 
\rowcolor{col_sur!10} TabPFN sur.\@ dir.\@ (v2.5-D, 8) & 6.501 & 1:31.9 && 7.277 & 3:35.5 && 0.449 & 9:26.8 && 0.245 & 19:46.8 && 0.088 & 7:16:12:43.1 \\ 
\rowcolor{col_sur!10} TabPFN sur.\@ dir.\@ (v2.5-R, 1) & 5.960 &   15.2 && 6.100 &   34.0 && 0.404 & 1:14.9 && 0.237 &  2:35.0 && 0.072 & 1:11:19:18.0 \\ 
\rowcolor{col_sur!10} TabPFN sur.\@ dir.\@ (v2.5-R, 8) & 5.074 & 1:34.3 && 5.802 & 3:32.0 && 0.445 & 9:19.2 && 0.229 & 19:08.8 && 0.062 & 7:22:01:05.3 \\ 

\rowcolor{col_sur!10} TabPFN sur.\@ aug.\@ (v2, 1)     & 1.282 &   13:00.5 && 1.298 &   28:42.4 && 0.166 &  2:08:16.4 && 0.176 &    4:55:53.6 && 0.033 & 1:18:05:05.4 \\ 
\rowcolor{col_sur!10} TabPFN sur.\@ aug.\@ (v2, 8)     & 1.235 & 1:18:37.4 && 1.247 & 2:48:38.6 && 0.153 & 11:41:32.9 && 0.151 & 1.01:02:50.8 &&   --- &          --- \\ 
\rowcolor{col_sur!10} TabPFN sur.\@ aug.\@ (v2.5-D, 1) & 1.201 &   22:33.5 && 1.233 &   23:56.9 && 0.178 &  1:44:08.8 && 0.187 &    6:19:12.1 && 0.031 & 1:14:00:12.6 \\ 
\rowcolor{col_sur!10} TabPFN sur.\@ aug.\@ (v2.5-D, 8) & 1.191 & 1:08:51.8 && 1.191 & 2:57:22.2 && 0.156 & 11:20:48.2 && 0.150 & 1:02:03:32.0 &&   --- &          --- \\ 
\rowcolor{col_sur!10} TabPFN sur.\@ aug.\@ (v2.5-R, 1) & 1.204 &   10:15.4 && 1.246 &   24:03.3 && 0.170 &  1:44:43.6 && 0.185 &    3:54:42.7 && 0.032 & 1:14:03:17.7 \\ 
\rowcolor{col_sur!10} TabPFN sur.\@ aug.\@ (v2.5-R, 8) & 1.193 & 1:10:12.0 && 1.193 & 2:46:02.4 && 0.154 & 11:25:14.6 && 0.145 & 1:02:15:57.0 &&   --- &          --- \\ 

\rowcolor{col_sur!10} TabPFN sur.\@ aug.\@ coal. (v2, 1)     & 1.194 &   12:27.2 && 1.230 &   30:50.7 && 0.186 &  1:58:51.6 && 0.174 &    5:03:48.4 && 0.032 & 2:00:27:09.3 \\ 
\rowcolor{col_sur!10} TabPFN sur.\@ aug.\@ coal. (v2, 8)     & 1.180 & 1:14:35.3 && 1.199 & 3:06:00.9 && 0.153 & 12:04:17.9 && 0.168 & 2:05:00:54.0 &&   --- &          --- \\ 
\rowcolor{col_sur!10} TabPFN sur.\@ aug.\@ coal. (v2.5-D, 1) & 1.182 &    9:36.7 && 1.211 &   23:30.2 && 0.182 &  1:29:03.2 && 0.185 &    3:59:28.3 && 0.032 & 1:12:47:43.9 \\ 
\rowcolor{col_sur!10} TabPFN sur.\@ aug.\@ coal. (v2.5-D, 8) & 1.163 & 1:04:15.7 && 1.175 & 2:38:26.5 && 0.155 & 10:24:36.5 && 0.148 & 2:10:26:25.7 &&   --- &          --- \\ 
\rowcolor{col_sur!10} TabPFN sur.\@ aug.\@ coal. (v2.5-R, 1) & 1.182 &    9:34.6 && 1.200 &   23:30.3 && 0.183 &  1:29:12.3 && 0.182 &    4:00:05.3 && 0.031 & 1:12:47:08.7 \\ 
\rowcolor{col_sur!10} TabPFN sur.\@ aug.\@ coal. (v2.5-R, 8) & 1.169 & 1:04:19.3 && 1.179 & 2:38:27.0 && 0.153 & 10:23:12.6 && 0.139 & 2:08:12:38.1 &&   --- &          --- \\ 

\specialrule{.8pt}{0pt}{6pt}
\end{tabular}
}
\end{table}

Table~\ref{tab:real_world_data} shows the $\operatorname{MSE}_{v}$ scores and computation times for the different methods across the five datasets. Separate regression with TabPFN performs the best on the first three datasets. For the $\texttt{Abalone}_\text{all}$ dataset, we observe a performance gain of $3.3\%$ over the previous best method (VAEAC) and a reduction in runtime from roughly 12 hours to 4 minutes (though CPU–GPU comparisons are not directly comparable). For the last two datasets, where the predictive function is non-smooth, TabPFN perform slightly worse than the tree-based regression methods. Like in the simulation study, this is expected and consistent with the recommendations by Olsen et al.~\cite{olsen2024comparative}, who suggest using non‑smooth models when explaining non‑smooth predictive functions. Nonetheless, TabPFN is not far behind on the \texttt{Adult} dataset and uses only a fraction of the computation time---only three days compared to 35 days.

We do not observe any consistent advantage of using a particular version of TabPFN among v2, v2.5-D and v2.5-R. For $\texttt{Abalone}_\text{cont}$, v2 achieves the best result whereas v2.5‑R matches or slightly outperforms the other versions on the remaining datasets. Regardless, these effects are quite small. Increasing the number of estimators from one to eight improves accuracy across nearly all settings, but at the expected cost of longer computation times. The increase in runtime is approximately linear in the number of forward passes.

As in the simulation studies, the surrogate TabPFN versions perform worse overall, but the TabPFN sur.~aug.~coal.~(v2.5‑D, 8) outperforms all non‑TabPFN methods on both versions of the \texttt{Abalone} data. The sur.~dir.~methods are among the weakest across all variants of TabPFN, highlighting the importance of including the missingness structure during in-context learning. We again observe that applying TabPFN separately for each coalition size (the aug.~coal.~methods) leads to consistently better performance than using a single model for all coalitions (the aug.~methods). This is likely because the coalition‑specific approach exposes each model to far more examples of the precise pattern of missing data it needs to predict for.

If TabPFN supported contexts larger than 50,000 instances, the performance of the surrogate regression approach might approach that of separate regression. This limitation is expected to diminish as future TabPFN versions increase the allowable context size (the enterprise version supports up to one million instances). A larger context would, however, increase prediction time, because TabPFN must condition on a greater number of instances. The augmented surrogate methods already have higher runtimes for this reason: their enriched contexts improve accuracy but make each forward pass more computationally expensive.

\section{Conclusion}
In this work, we have proposed the use of TabPFN for estimating conditional Shapley values. Tabular foundation models are rapidly gaining traction, with several prominent architectures such as TabICL, AutoGluon, MITRA and TabNet under active development \cite{arik2021tabnet, erickson2020autogluon, qu2025tabicl, zhang2025mitra}. We have focused on TabPFN due to its strong empirical performance on the \href{https://huggingface.co/spaces/TabArena/leaderboard}{TabArena} leaderboards and its widespread recognition in the machine learning community.

When combined with separate regression, TabPFN consistently performs well and often outperforms existing state-of-the-art Monte Carlo-based methods and regression‑based approaches, particularly when the predictive model is smooth. As expected, performance declines somewhat for non‑smooth predictive functions, but TabPFN still performs almost as well as tree-based methods while requiring far less computation. TabPFN performs worse in the surrogate regression formulation, probably because the synthetic pre‑training data did not adequately include the censoring patterns typical of Shapley value regression, with entire columns of data missing. A promising direction for future work is pre-training or fine-tuning TabPFN or other tabular foundation models on datasets which explicitly include such patterns of missing data. This is especially important for the surrogate approach, where the model only needs to be fine-tuned once.

Overall, our findings indicate that TabPFN is a promising and competitive tool for Shapley value estimation, particularly when used in the separate formulation, and its potential is likely to grow as tabular foundation models continue to advance.

\begin{credits}
\subsubsection{\ackname} The work by Lars H.\ B.\ Olsen is funded by the Norwegian Research Council’s Center of Excellence Integreat (project no.\ 332645).

\subsubsection{\discintname}
The authors have no competing interests to declare that are
relevant to the content of this article.
\end{credits}

%
%
%
\bibliographystyle{splncs04}
\bibliography{bibliography}

\end{document}